\documentclass[journal]{IEEEtran}
\usepackage{cite}

\ifCLASSINFOpdf
\else
\fi

\usepackage{amsmath}
\usepackage{amssymb}
\usepackage{commath}
\usepackage{graphicx}
\usepackage{multirow}
\usepackage{caption}
\usepackage{subcaption}
\usepackage{hyperref}
\usepackage{scalerel}
\usepackage{tikz}

\usetikzlibrary{svg.path}
\definecolor{orcidlogocol}{HTML}{A6CE39}
\tikzset{
  orcidlogo/.pic={
    \fill[orcidlogocol] svg{M256,128c0,70.7-57.3,128-128,128C57.3,256,0,198.7,0,128C0,57.3,57.3,0,128,0C198.7,0,256,57.3,256,128z};
    \fill[white] svg{M86.3,186.2H70.9V79.1h15.4v48.4V186.2z}
                 svg{M108.9,79.1h41.6c39.6,0,57,28.3,57,53.6c0,27.5-21.5,53.6-56.8,53.6h-41.8V79.1z M124.3,172.4h24.5c34.9,0,42.9-26.5,42.9-39.7c0-21.5-13.7-39.7-43.7-39.7h-23.7V172.4z}
                 svg{M88.7,56.8c0,5.5-4.5,10.1-10.1,10.1c-5.6,0-10.1-4.6-10.1-10.1c0-5.6,4.5-10.1,10.1-10.1C84.2,46.7,88.7,51.3,88.7,56.8z};
  }
}

\newcommand\orcidicon[1]{\href{https://orcid.org/#1}{\mbox{\scalerel*{
\begin{tikzpicture}[yscale=-1,transform shape]
\pic{orcidlogo};
\end{tikzpicture}
}{|}}}}

\usepackage{hyperref} 

\begin{document}
%
\title{

Exploring Deep 3D Spatial Encodings for Large-Scale 3D Scene Understanding
}

%
%
%

\author{Saqib~Ali~Khan~\orcidicon{0000-0002-8967-7442} Student Member, IEEE,
        Yilei~Shi~\orcidicon{0000-0003-1907-8214},~\IEEEmembership{Member,~IEEE,}
        Muhammad~Shahzad~\orcidicon{0000-0002-8278-9118},~\IEEEmembership{Member,~IEEE,}
        Xiao~Xiang~Zhu~\orcidicon{0000-0001-5530-3613}\,~\IEEEmembership{Senior Member,~IEEE}

\thanks{This work was supported through Deutsch-Pakistanische Forschungskooperationen (grant number: 57459164) by the German Academic Exchange Service (DAAD). 2j ab 19}
\thanks{S.A. Khan and M. Shahzad are with the School of Electrical Engineering and Computer Science (SEECS), National University of Sciences and Technology (NUST), Islamabad, Pakistan (email: sakhan.bscs16seecs@seecs.edu.pk;muhammad.shehzad@seecs.edu.pk)}
\thanks{Y. Shi is with the Chair of Remote Sensing Technology (LMF), Technical University of Munich (TUM), 80333 Munich, Germany and X. X. Zhu is with Signal Processing in Earth Observation (SiPEO), Technical University of Munich (TUM), 80333 Munich, Germany, and also with the Remote Sensing Technology Institute (IMF), German Aerospace Center (DLR), 82234 Wessling,
Germany (e-mail: xiaoxiang.zhu@dlr.de; yilei.shi@tum.de)}
}

%
%

\markboth{Submitted to IEEE Geoscience and Remote Sensing Letters, October~2020}%
{Shell \MakeLowercase{\textit{et al.}}: Bare Demo of IEEEtran.cls for Journals}
%



\maketitle

\begin{abstract}

Semantic segmentation of raw 3D point clouds is an essential component in 3D scene analysis, but it poses several challenges, primarily due to the non-Euclidean nature of 3D point clouds. Although, several deep learning based approaches have been proposed to address this task, but almost all of them emphasized on using the latent (global) feature representations from traditional convolutional neural networks (CNN), resulting in severe loss of spatial information, thus failing to model the geometry of the underlying 3D objects, that plays an important role in remote sensing 3D scenes. In this letter, we have proposed an alternative approach to overcome the limitations of CNN based approaches by encoding the spatial features of raw 3D point clouds into undirected symmetrical graph models. These encodings are then combined with a high-dimensional feature vector extracted from a traditional CNN into a localized graph convolution operator that outputs the required 3D segmentation map. We have performed experiments on two standard benchmark datasets (including an outdoor aerial remote sensing dataset and an indoor synthetic dataset). The proposed method achieves on par state-of-the-art accuracy with improved training time and model stability thus indicating strong potential for further research towards a generalized state-of-the-art method for 3D scene understanding.

\end{abstract}


%
\IEEEpeerreviewmaketitle

\section{Introduction}
%
%
%
%

\IEEEPARstart{R}{apid} advances in consumer-grade 3D acquisition sensors, such as RGB-D cameras and LIDARs have enabled the reconstruction of high quality 3D scenes, which has led to an increased demand for faster and more reliable scene classification methods. The outcome of such sensors is usually a 3D point cloud providing the geometrical information of the underlying 3D scene. Semantic segmentation of these point clouds refers to assigning each point a certain known label and is one of most the fundamental task in the context of scene interpretation. However, the semantic segmentation of raw 3D point clouds remains a highly challenging task owing to the lack of implicit neighborhood relationships that exists between the points in 3D space. Furthermore, the point clouds do not posses regular geometries and lie in a non-Euclidean space, so there is not a single well-defined method that can enable convolution on raw 3D point clouds. Owing to the sparse irregularity of 3D point clouds, several researchers \cite{Kumawat2019LP3D, Qi2016VolumetricAM, multi-view-cnn} have used a preprocessing step of voxelization prior to applying convolution, but doing this step adds another layer of expensive computation and the transformed structure loses most of the spatial arrangement of points, and therefore the network struggles to propagate the spatial features that can improve the overall segmentation results.

So, in order to encode the spatial features more effectively, several researchers have used graph representations for non-Euclidean datasets. For instance, Bronstein et al. \cite{geometric}, first coined the term geometric deep learning as an umbrella term for methods attempting to generalize deep neural networks on datasets that can be represented on a graph. However, Bruna et al. \cite{bruna} proposed the first evidence of the generalization of graph convolutional networks in spectral domain, but without taking 3D translational factors into consideration. Moreover, Defferrard et al. \cite{defferrard} proposed a fast convolution that uses the Chebyshev polynomials recursively for high-dimensional non-Euclidean spectral graphs such as protein interaction networks or social networks. Furthermore, few researchers prefer the kernel-based approaches \cite{kernel1, kernel2}, which is based on encoding the local spatial features using graph kernels that contain meaningful information which can greatly improve the classification results but may suffer from quadratic training complexity and higher memory footprint.

Fortunately, with the increasing use of convolutional neural networks (CNNs) in 2D image analysis and high availability of hardware resources (GPUs), there is a rapidly growing interest to directly process raw 3D point clouds using traditional CNNs \cite{Kumawat2019LP3D, pointnet, pointet++, splatnet}. But, unlike 2D images, that lie in a finite Euclidean domain, 3D point clouds do not posses regular geometries and thus do not lie on a regular grid. It is because of this property, traditional CNNs fail to generalize well on objects at varying contextual scales. Furthermore, CNN-based methods overlook the pre-existing neighbourhood information inside the points in 3D space, that can be very beneficial when classifying objects at scale.

\begin{figure*}
\begin{center}
\includegraphics[width=0.95\linewidth]{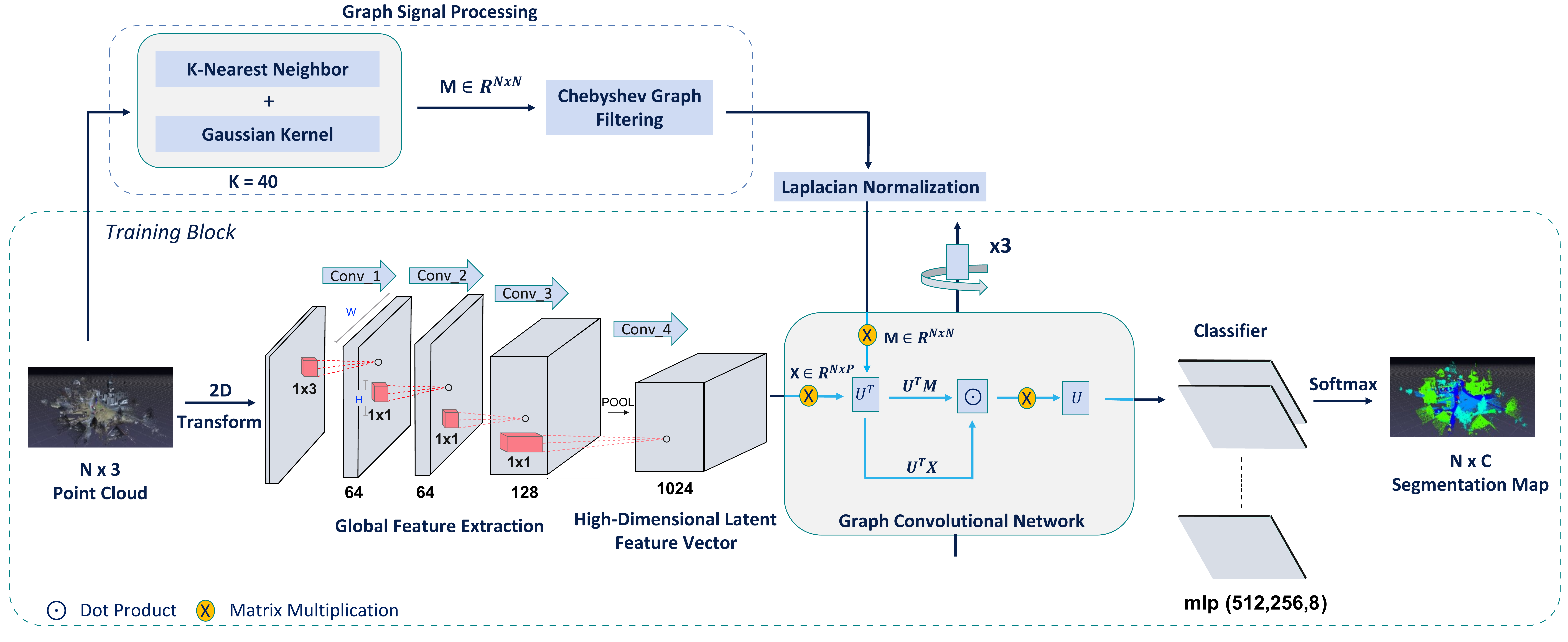}
\end{center}
   \caption{\small{ Our proposed network architecture: The figure shows the building blocks and layer configurations of our proposed semantic segmentation network. The graph signal processing is a preprocessing step and is only carried out once.}  }
\label{fig:architecture}
\end{figure*}

Recent studies predict that deep learning has proven to be a major breakthrough in many fields. In particular, there exists a number of challenges for using deep learning in remote sensing domain \cite{remote-sensing-survey}. In this letter, we kept in mind the complexity of structural and spatial patterns in remote sensing scenes and proposed a network that jointly uses both global (CNN-based) and local features into a unified architecture for semantic segmentation of 3D remote sensing scenes. Our network encodes the raw 3D point clouds into undirected symmetrical graphs using the k-Nearest Neighbor algorithm, thus preserving the spatial information. In our proposed 3D graph representation, we have only used the 3D Cartesian coordinates $[x, y, z]$ to define a single point. After our feature encoding step, we have used 2D convolutional layers inspired from PointNet \cite{pointnet}, to extract a high-dimensional feature vector that summarizes the latent (global) information and provides geometric invariance \cite{pointnet}. Lastly, we combine the outputs of our graph encoder as well as 2D feature extractor into a graph convolutional network (GCN). Spatio-temporal graph neural network \cite{survey} or GCN, refines the latent feature representations using the graph embedded spatial features to output a global template, that summarizes each point inside the symmetrical graph. Using the spatial features, our network learns the object's structure encoded in the graph and achieves a much faster convergence rate compared to state-of-the-art models (see section III-C for details). We have performed experiments on the aerial Semantic3D and the indoor S3DIS benchmark datasets. The first evaluation using standard metrics shows that the proposed method achieves on par state-of-the-art performance. Our main contributions are summarized as follows:

\begin{itemize}
    \item We propose a novel end-to-end architecture for semantic segmentation of remote sensing 3D scenes that directly consumes raw 3D point clouds;
    \item We combine the 3D latent features with spatial features in a non-trivial manner for increased performance and geometric invariance.
    \item We provide the results of our first evaluation on two benchmarking datasets on which our model shows dominant performance compared to most state-of-the-art models with faster convergence rate.
\end{itemize}

\section{Proposed Network Architecture}

\subsection{Graph Feature Encoder} 
\begin{figure}
\begin{center}
\includegraphics[width=1.0\linewidth]{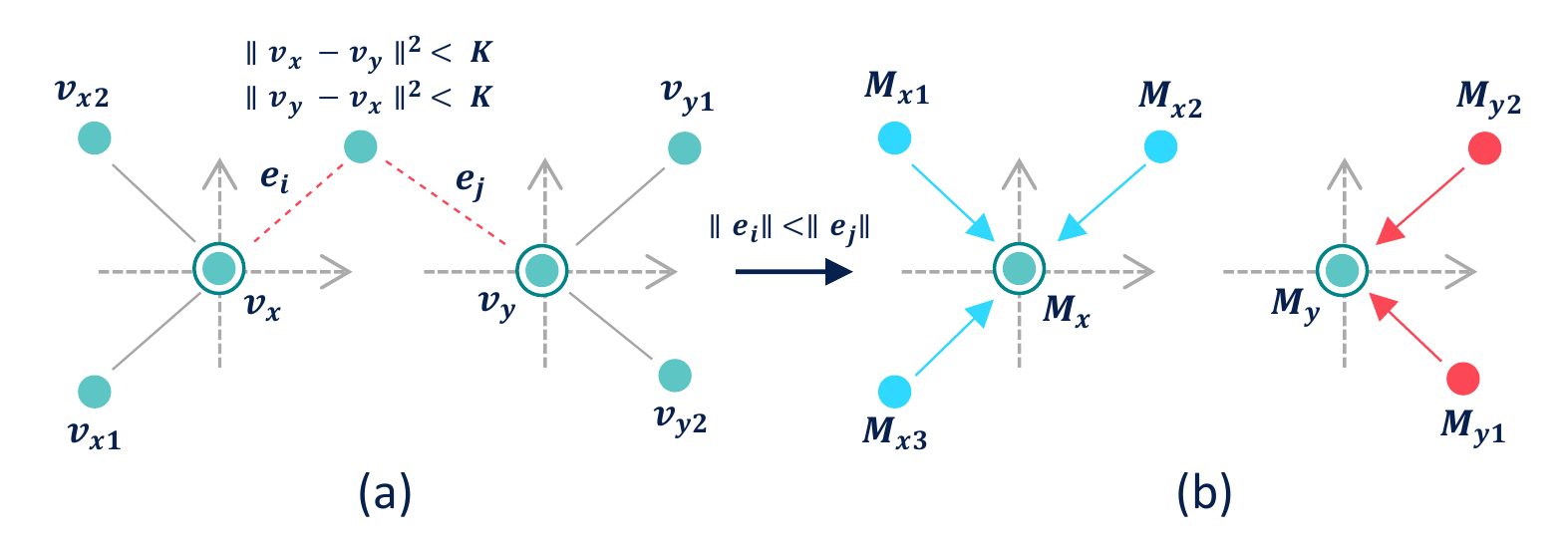}
\end{center}
   \caption{\small{Point cloud transformation to weighted graph signals using k-NN with K = 40. Each edge is labelled with the Euclidean distance between the two vertices.} }
\label{fig:graph_construction}
\end{figure}

A graph convolutional network (GCN) takes as input a graph $G$, with a finite number of vertices or nodes $v_x \in V$, and edges $e_{xy} \in E$, where $e_{xy} = \|v_x, v_y\|$ representing a neighbourhood connection between the vertices $v_x$ and $v_y$. The graph edges or the spatial connections between the points are represented as a weighted graph signal $M_{xy}$ that represents an entry $M_{x,y}$ into the adjacency matrix $M$, encoding a spatial connection between $v_x$ and $v_y$.  Additionally, in order to calculate the value of $M_{x,y}$, we use the k-Nearest Neighbor algorithm, where k is a hyper-parameter and it represents the number of neighbors we calculate for each point based on the distance as illustrated in Fig \ref{fig:graph_construction}. Furthermore, we use a Gaussian kernel to weigh the edges $e_{xy}$ of vertices $v_x$ and $v_y$, as given by equation \ref{gaussian}.

\begin{equation}
M_{x,y} = 
\left\{
\begin{array}{ll}
      \exp(- \frac{\|v_x-v_y\|^2}{2\sigma^2}) & \textrm{if }  \|v_x-v_y\| < \kappa \\
      0 & \textrm{otherwise} \\
\end{array}
\right. 
\label{gaussian}
\end{equation}

where $\sigma$ represents the standard deviation, and $\kappa$ is the distance threshold on $\|v_x-v_y\|$, representing the Euclidean distance between the vertices $v_x$ and $v_y$.

Given the adjacency matrix $M \in \mathbb{R}^{N \times N}$, we use graph filtering techniques \cite{kipf, graph_class} for vertex-domain transformations. Specifically, we have used a normalized Laplacian matrix given by, $L(x) = I_n - D^{-\frac{1}{2}}MD^{-\frac{1}{2}}$, where D represents the diagonal matrix of $M$ such that $D_{x,y}=\Sigma_y[M_{x,y}]$. The $g(x)$ or the graph mapping function is a linear transformation function with inputs $x$ and coefficients $\alpha_1, \alpha_2,....,\alpha_n$,

\begin{equation}
    g(x) = f_{\alpha}(L)x = \sum_{i=0}^{K} \alpha_i L^i (x)
\end{equation}

In spectral-based image analysis, some techniques \cite{wavelets, chebnet, graph_signal_processing} employ Chebyshev polynomials in order to approximate the graph filter coefficients e.g. ChebyNet \cite{chebnet} uses the eigen decomposition of Laplacian with diagonal matrix $L(x)$ as,

\begin{equation}
    f(x) = g_{\theta}(L)x = \sum_{i=0}^{K}\theta_iT_i(L)x
\label{chebyshev}
\end{equation}

where the goal of our GCN is to find the optimal values of the graph filtering  coefficients $\{\theta,\alpha\}$ by using the Chebyshev graph filter approximation given by equation \ref{chebyshev}. Furthermore, Defferrard et al. \cite{defferrard}, provides evidence that using ChebyNet \cite{chebnet} in homogeneous graphs can prove to be very effective in 2D image analysis. Therefore, we have adopted a similar approach to ChebyNet \cite{chebnet}, but we apply the method to heterogeneous graphs i.e. non-Euclidean 3D point clouds.

\subsection{Global Feature Extractor}

The unreasonable effectiveness of convolutional neural networks (CNNs) in 2D image analysis has motivated researchers to apply the same techniques to process raw 3D point clouds. PointNet \cite{pointnet}
is one of the pioneer research that directly consumes raw 3D point clouds to extract the latent feature representation that is further classified into semantic labels. Furthermore, PointNet \cite{pointnet} uses a stack of 2D convolutional layers in a way that ensures invariance to geometric transformations and point cloud permutations. At that time, PointNet \cite{pointnet} outperformed all state-of-the-art approaches, mostly based on point cloud transformations.

In this letter, we have used an approach that is inspired by PointNet \cite{pointnet} to extract high order latent features prior to using GCN. This additional step adds more robustness to our architecture both in terms of performance (See section III for details) and reliability i.e. the model is invariant to permutations and geometric transformations. In order to perform this step, we have used a block of four 2D CNN layers of depth $[64,64,128,1024]$, that acts as global feature extractor and is passed as a template to the graph convolutional network. So, our GCN takes as input a high-dimensional feature vector $\{x_i^{(1)}, x_i^{(2)}, .... x_i^{(P)}\} \in \mathbb{R}^{N \times P}$, where $P$ represents the cardinality of the feature vector, instead of using the typical 3D cartesian coordinates $\{x,y,z\}$ \cite{gcn_sim}.

\subsection{Graph Convolutional Network}

In our proposed methodology, the GCN takes as input the weighted graph signals $M \in \mathbb{R}^{N \times N}$ and a feature vector $X \in \mathbb{R}^{N \times P}$, and its goal is to find the optimal values of graph filtering coefficients given by equation \ref{chebyshev}. Furthermore, our GCN is defined as a non-linear translation function $ \delta $, that maps the graph nodes or points to propagate the spatial information corresponding to their relative positions in 3D space. An $l$-th GCN layer can be written as in equation \ref{mapper},

\begin{equation}
    f(X^{(l)}, M) = \delta \hspace{2pt}(X^{(l)} \theta^{(l)} M))
\label{mapper}
\end{equation}

where $\theta$ corresponds to the learnable parameter of order $K$. Moreover, our graph $M$ can contain values of varying range, and therefore, our GCN cannot generalize effectively on graphs in varying spectral domains. So, we have used graph Laplacian for the purpose of normalization  $L(x) = I_n - D^{-\frac{1}{2}}MD^{-\frac{1}{2}}$, where $D$ represents the diagonal matrix of $M$ such that $D_{ii}=\sum_j M_{ij}$. Intuitively, in symmetric normalization, we add all the rows of graph $M$ such that it always sums to one \cite{kipf}. So, equation \ref{mapper} becomes,

\begin{equation}
    f(X^{(l)}, M) = \delta \hspace{2pt} (X^{(l)} \theta^{(l)}  \widehat{D}^{-\frac{1}{2}} \widehat{M} \widehat{D}^{-\frac{1}{2}}   ))
\end{equation}

where $\widehat{M} = M + I$ and $I$ is the identity matrix. Additionally, we can use the inverse graph Fourier transform \cite{survey}, to define a graph convolutional operator using the matrix of eigenvectors $U^{T}m$ of the input graph signal $m$ and $U^{T}x$ of the feature vector $x$,

\begin{equation}
    x*_Gm = U(U^Tx \odot U^Tm)
\end{equation}

where $\odot$ represents pointwise product. In the proposed method, we used three layers of graph convolutions with Chebyshev order $K=3$, being a hyperparameter. The proposed architecture diagram can be visualized in Fig \ref{fig:architecture}.

\section{Experiments}
There is a long tradition of evaluating deep learning architectures on geospatial datasets, particularly ISPRS.  The ISPRS \textit{Benchmark on Urban Object Detection and Reconstruction}\footnote{https://www2.isprs.org/commissions/comm2/wg4/benchmark/detection-and-reconstruction/}, offers many challenges in the geospatial domain, including 3D reconstruction, object detection and 2D/3D semantic segmentation. The outcome of these challenges are applied to real world remote sensing problems like building segmentation \cite{building-segmentation} in large metropolitan areas. In this letter, we have evaluated our architecture on Semantic3D benchmark dataset that contains outdoor scenes from various urban and rural environments.

\begin{figure}
     \centering
     \begin{subfigure}[b]{0.24\textwidth}
         \centering
         \includegraphics[width=\textwidth]{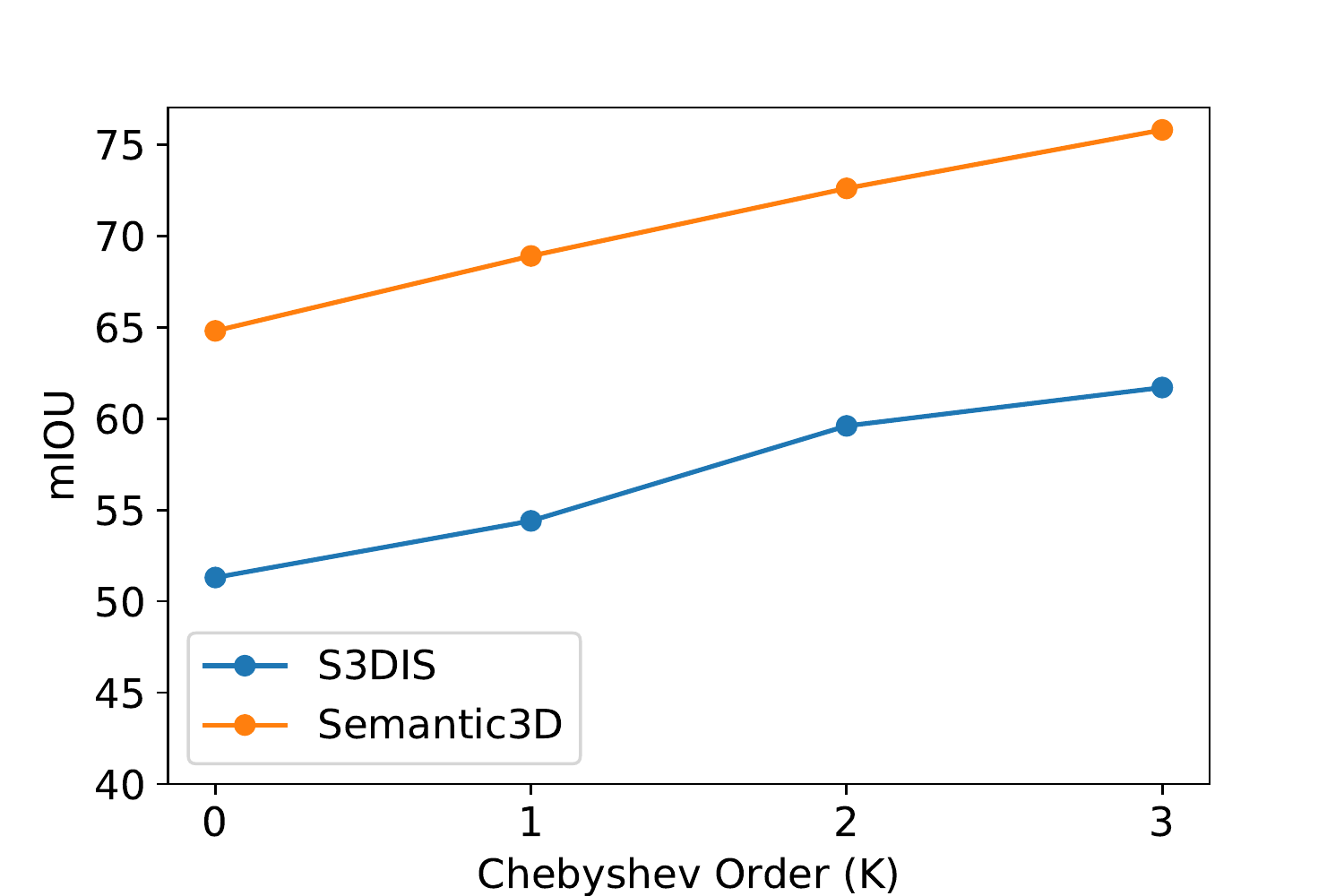}
         \caption{ }
         \label{fig:cheb_order}
     \end{subfigure}
     \begin{subfigure}[b]{0.24\textwidth}
         \centering
         \includegraphics[width=\textwidth]{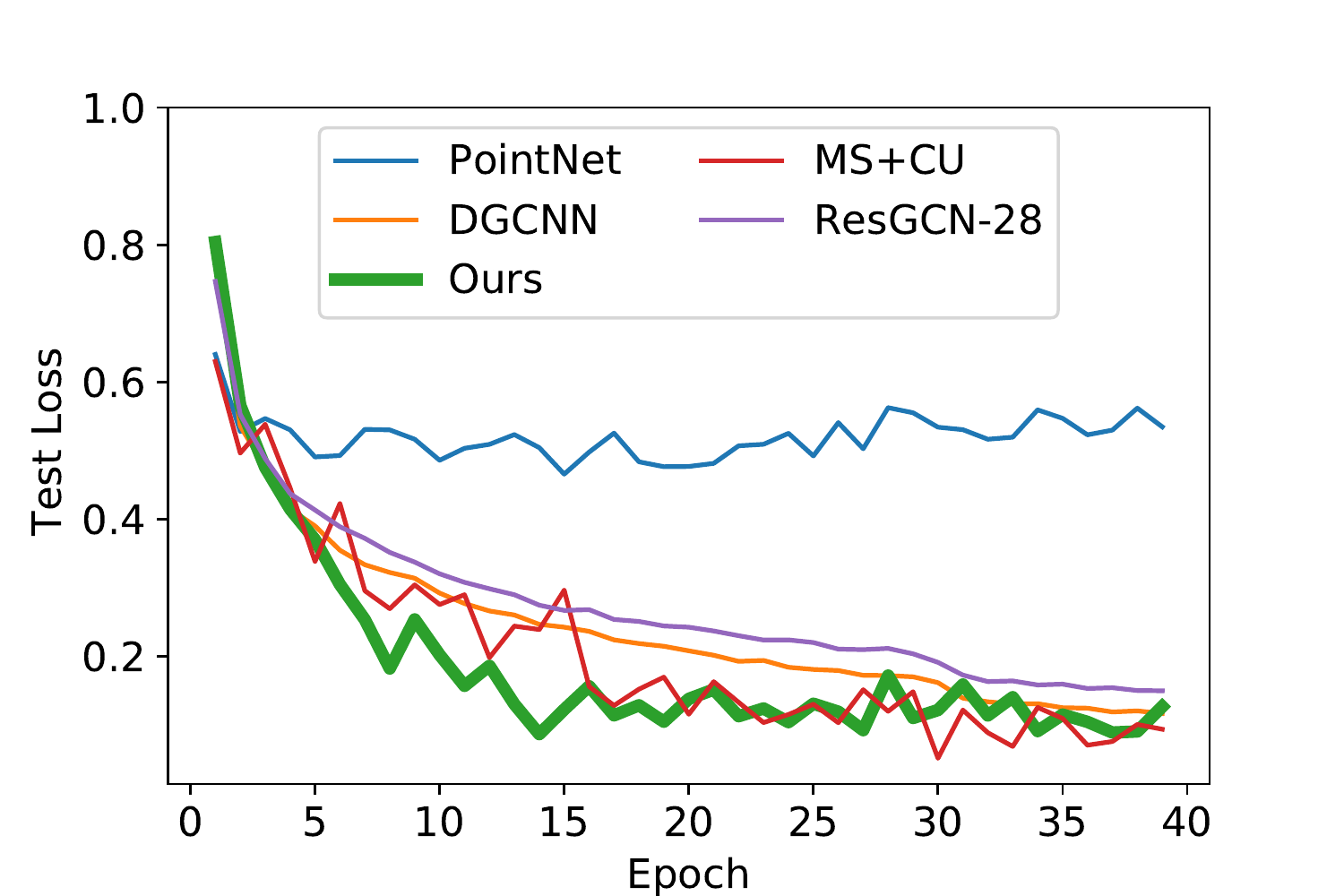}
         \caption{ }
         \label{fig:test_loss}
     \end{subfigure}
        \caption{\small{Performance Analysis based on, (a) The effect of using different Chebyshev order $K$ on the overall mIOU, and (b) The convergence rate using test loss on S3DIS dataset.  }}
        \label{fig:comparison_graphs}
\end{figure}

\subsection{Datasets and Comparisons}
In our experiments, we have used two benchmarking datasets for indoor and outdoor scenes and the results are shown in Table \ref{results}.

\subsubsection{S3DIS} We have benchmarked our method on the Stanford 3D dataset \cite{s3dis} that contains $271$ indoor spaces in $6$ different areas. The point clouds are divided into $1$m by $1$m blocks in a same way as in \cite{pointnet}. The dataset contains $13$ semantic categories and $9$ different features per point, but we only use $3$ i.e. $(x,y,z)$ coordinates for simplicity.

\subsubsection{Semantic3D} Semantic3D \cite{semantic3d} is by the far the largest labelled 3D dataset containing point clouds from outdoor scenes, including both rural and urban environments. This dataset contains nearly $4$ billion points including most of the scenes across central Europe collected with $30$ terrestrial laser scanners. We have used the \textit{reduced-8} version of this dataset for training in our experiments, which has $8$ different categories. Furthermore, we take the mean of intersection-over-union (IOU) of all the $8$ classes in a similar way as defined in \cite{semantic3d} as our benchmarking metric. 


\subsection{Implementation Details}
The proposed architecture is trained using an Adam optimizer with a batch size of $32$ and a learning rate of $1e^{-3}$. In this letter, we preprocess the datasets by dividing the point clouds into blocks of $1m$ by $1m$ area, with $4096$ points per block. Furthermore, in order to avoid overfitting, we have used dropout regularization of $0.4$ for CNN and $0.8$ for GCN blocks, and a weight decay of $2e^{-4}$ to avoid complexity.

In order to train the network, we have used $2$x NVIDIA Titan X GPUs with $64$ GB on-board CPU memory for preprocessing and visualizations. However, the training time and memory consumption greatly depends on the order of Chebyshev polynomial $K$ given by equation \ref{chebyshev}; With increasing order of K, the size of $T_{i}L$ increases with increase in sparsity and so does the overall performance of the network, but with quickly diminishing rates, as evident by Fig \ref{fig:comparison_graphs}(a). Additionally, increasing the order of $K$, increases the overall training time and computational footprint. In this letter, we have trained our network on $K=3$ for maximum performance and the total training time is $96$ hours.

\begin{figure*}
\begin{center}
\includegraphics[width=1.0\textwidth]{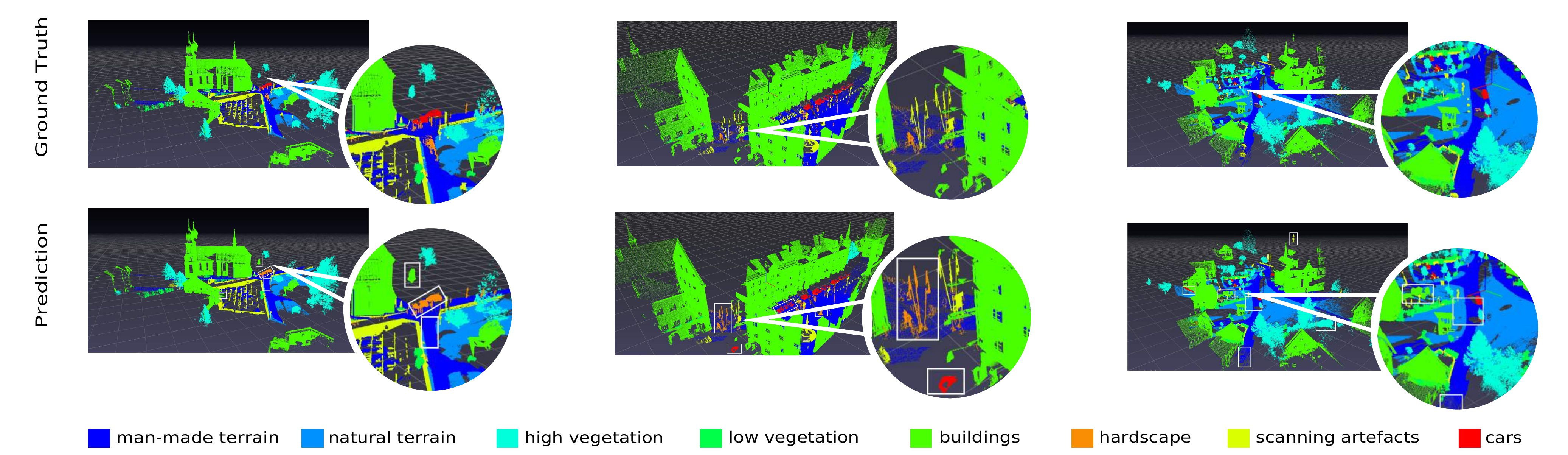}
\end{center}
   \caption{\small{\textbf{Qualitative analysis of evaluation on Semantic3D dataset}: We have used the open source PPTK viewer for point cloud visualization and Open3D to convert the sparse output of the network to dense 3D point cloud using k-NN hybrid search with k=20 and radius=0.2. The classification errors are highlighted for better visualization. (Best viewed in color) } }
\label{fig:qualitative_results}
\end{figure*}

\subsection{Performance Analysis}
To further explain the architecture design goals, we investigate the roles of local (spatial) and global features in a unified architecture:

\begin{table}
\begin{center}
\caption{\textbf{Quantitative results on S3DIS and Semantic3D benchmark datasets:} The mean IOU and accuracy is calculated as an average of IOUs and accuracies of all categories.} 
\label{results}
\begin{tabular}{c|l|cc}
\hline
&\textbf{Method} & \textbf{mean IOU} & \textbf{mean Accuracy} \\
\hline
& Pointnet \cite{pointnet} & 47.6 & 78.5 \\\cline{2-4} 
& MS+CU \cite{grcu} & 49.7 & 81.1 \\\cline{2-4}
& DGCNN \cite{dgcnn} & 56.1 & 84.1 \\\cline{2-4}
\multirow{2}{*}{\textbf{S3DIS}}& Pointnet++ \cite{pointet++} & 53.2 & -\\ \cline{2-4}
& ResGCN-28 \cite{deepgcns} & 60.0 & 85.9\\\cline{2-4}

\cline{2-4}
&\textbf{Ours (GCN Only)} & 60.6 & 83.6 \\ \cline{2-4}
&\textbf{Ours} & \textbf{61.7} & \textbf{86.3} \\ \hline

\hline
& & & \\
\hline
& RandLA-Net \cite{randla} & \textbf{0.774} & 0.948 \\ \cline{2-4}
&RGNet \cite{rgnet} & 0.747 & 0.945 \\ \cline{2-4}
\multirow{2}{*}{\textbf{Semantic3D}} &KPConv \cite{kpconv} & 0.746 & 0.929\\ \cline{2-4}
&SPGraph \cite{spgraph} & 0.732 & 0.940\\
\cline{2-4}
&\textbf{Ours (GCN Only)} & 0.745 & 0.922 \\ \cline{2-4}
&\textbf{Ours} & 0.758 & \textbf{0.951} \\ \hline

\end{tabular}
\end{center}
\end{table}

\subsubsection{Effect of using spatial features} The traditional CNNs can prove to be very effective for Euclidean datasets, but many of the CNN based approaches \cite{Qi2016VolumetricAM, pointnet, pointet++} applied on 3D point clouds try hard to learn the spatial information of the underlying 3D objects in the illuminated scenes. It is because the 3D point clouds are unstructured, and they do not posses regular geometries, therefore, we cannot define a single convolution operation that can learn the underlying neighbourhood features for point clouds with $N$! geometric permutations. Consider Fig \ref{fig:comparison_graphs}(b), the issue is more severe for Pointnet \cite{pointnet}, which uses the global representations of 2D CNNs in the context of 3D classification. On the other hand, our network is much more stable on the non-Euclidean 3D point clouds and we achieve a much faster convergence rate. Furthermore, using the spatial information makes our network more robust at identifying object at varying scales.

\subsubsection{Effect of using global features} Using only the spatial features is not enough, because there can be N! possible permutations of 3D objects and the GCN alone is not invariant to all permutations. For this reason, we extract a high dimensional feature vector containing embeddings that serve as a template for graph convolutional network. This adds extra precision and reliability to our network across different scenes with varying object geometries and scales (Refer to Fig \ref{fig:comparison_graphs}(b)). Additionally, our network preserves the spatial information across the entire network because we encode this information into graphs that are symmetric by nature.

The final results are shown in Table \ref{results}, and it is evident that our network achieves on par state-of-the-art performance compared to existing architectures with added spatial features and the performance is further increased by using latent features. Furthermore, the results of testing on Semantic3D benchmark dataset can be visualized in Fig \ref{fig:qualitative_results}. There are some classification errors for smaller non-obvious objects which are highlighted in the figure.

\section{Conclusion}
In this letter, an end-to-end deep-feature based graph convolutional network that preserves the spatial geometry of 3D objects has been presented. The proposed method of embedding the point clouds using graph representation highlights the importance of using spatial features for the task of 3D semantic segmentation. Moreover, the proposed network uses traditional CNNs to extract 3D latent features that serves as a meaningful template for the GCN and also provides geometric invariance which in turn increases the reliability of our network across a variety of scenes. The first evaluation has shown promising potential on indoor and outdoor scenes from two standard benchmarking datasets, but the proposed network has high computational footprint due to using static graphs. In the future work, we intend to optimize the static graph conversion process by using the dynamic subgraphs during the training process and will also incorporate a separate module that will propagate the latent (global) features between the subgraphs, potentially improving the overall classification results. 



%




\ifCLASSOPTIONcaptionsoff
  \newpage
\fi



\bibliographystyle{IEEEtran}
\bibliography{IEEE}
\end{document}